\documentclass[sigconf]{acmart}
\usepackage{multirow}
\usepackage{makecell}
\usepackage{subfigure}
\usepackage{multicol}
\usepackage{multirow}
\usepackage{url}
\usepackage{amsmath,amsfonts,amsthm} 
\usepackage{graphicx}
\usepackage{subfigure}
 \usepackage{balance}
\usepackage{color}
\usepackage{tabularx}

\AtBeginDocument{%
  \providecommand\BibTeX{{%
    \normalfont B\kern-0.5em{\scshape i\kern-0.25em b}\kern-0.8em\TeX}}}

\setcopyright{acmcopyright}
\copyrightyear{2021}
\acmYear{2021}
\acmDOI{10.1145/1122445.1122456}

\acmConference[SIGIR '21]{SIGIR '21:In Proceedings of the International ACM SIGIR Conference
on Research and Development in Information Retrieval}{June 03--05, 2021}{Woodstock, NY}
\acmBooktitle{SIGIR '21:  ACM SIGIR Conference
on Research and Development in Information Retrieval,
  June 03--05, 2021, Woodstock, NY}
\acmPrice{15.00}
\acmISBN{978-1-4503-XXXX-X/18/06}

\begin{document}

\title{A Graph-guided Multi-round Retrieval Method for Conversational Open-domain Question Answering}


\author{Yongqi Li}\affiliation{The Hong Kong Polytechnic University\country{China}}\email{liyongqi0@gmail.com}
\author{Wenjie Li}\affiliation{The Hong Kong Polytechnic University\country{China}}\email{cswjli@comp.polyu.edu.hk}
\author{Liqiang Nie}\affiliation{Shandong University\country{China}}\email{nieliqiang@gmail.com}

\renewcommand{\shortauthors}{Anonymous Authors.}

\begin{abstract}
In recent years, conversational agents have provided a natural and convenient access to useful information in people’s daily life, along with a broad and new research topic, conversational question answering (QA). Among the popular conversational QA tasks, conversational open-domain QA, which requires to retrieve relevant passages from the Web to extract exact answers, is more practical but less studied. The main challenge is how to well capture and fully explore the historical context in conversation to facilitate effective large-scale retrieval. The current work mainly utilizes history questions to refine the current question or to enhance its representation, yet the relations between history answers and the current answer in a conversation, which is also critical to the task, are totally neglected. To address this problem, we propose a novel graph-guided retrieval method to model the relations among answers across conversation turns. In particular, it utilizes a passage graph derived from the hyperlink-connected passages that contains history answers and potential current answers, to retrieve more relevant passages for subsequent answer extraction. Moreover, in order to collect more complementary information in the historical context, we also propose to incorporate the multi-round relevance feedback technique to explore the impact of the retrieval context on current question understanding. Experimental results on the public dataset verify the effectiveness of our proposed method. Notably, the F1 score is improved by 5\% and 11\% with predicted history answers and true history answers, respectively.  
\end{abstract}




\maketitle

\section{Introduction}
In recent years, the rise of machine learning techniques has accelerated the development of conversational agents, such as Alexa\footnote{\url{https://www.alexa.com/}.}, Siri\footnote{\url{https://www.apple.com/siri/}.}, and Xiaodu\footnote{\url{https://dueros.baidu.com/}.}. These conversational agents provide a natural and convenient way for people to chit-chat, complete well-specified tasks, and seek information in their daily life. People often prefer to ask conversational questions when they have complex information needs or are of interest to certain broad topics. It is therefore essential to endow conversational agents with the capability to answer conversational questions, which introduces a broad and new research area, namely conversational question answering (QA). Nowadays, conversational QA has attracted more and more researchers who have developed various tasks with different emphases, including but not limited to conversational knowledge-based QA~\cite{christmann2019look,zhang2020summarizing}, conversational machine reading comprehension~\cite{reddy2019coqa}, conversational search~\cite{hashemi2020guided}, and conversational open-domain QA~\cite{Qu0CQCI20}. Among them, conversational open-domain QA is more practical yet more challenging. Apart from highly historical context-dependent, elliptical, and even unanswerable questions, it requires to retrieve the relevant passages from the Web and to extract the text spans from the retrieved passages as the exact answers to the given questions.

\begin{figure}

  \includegraphics[width=\linewidth]{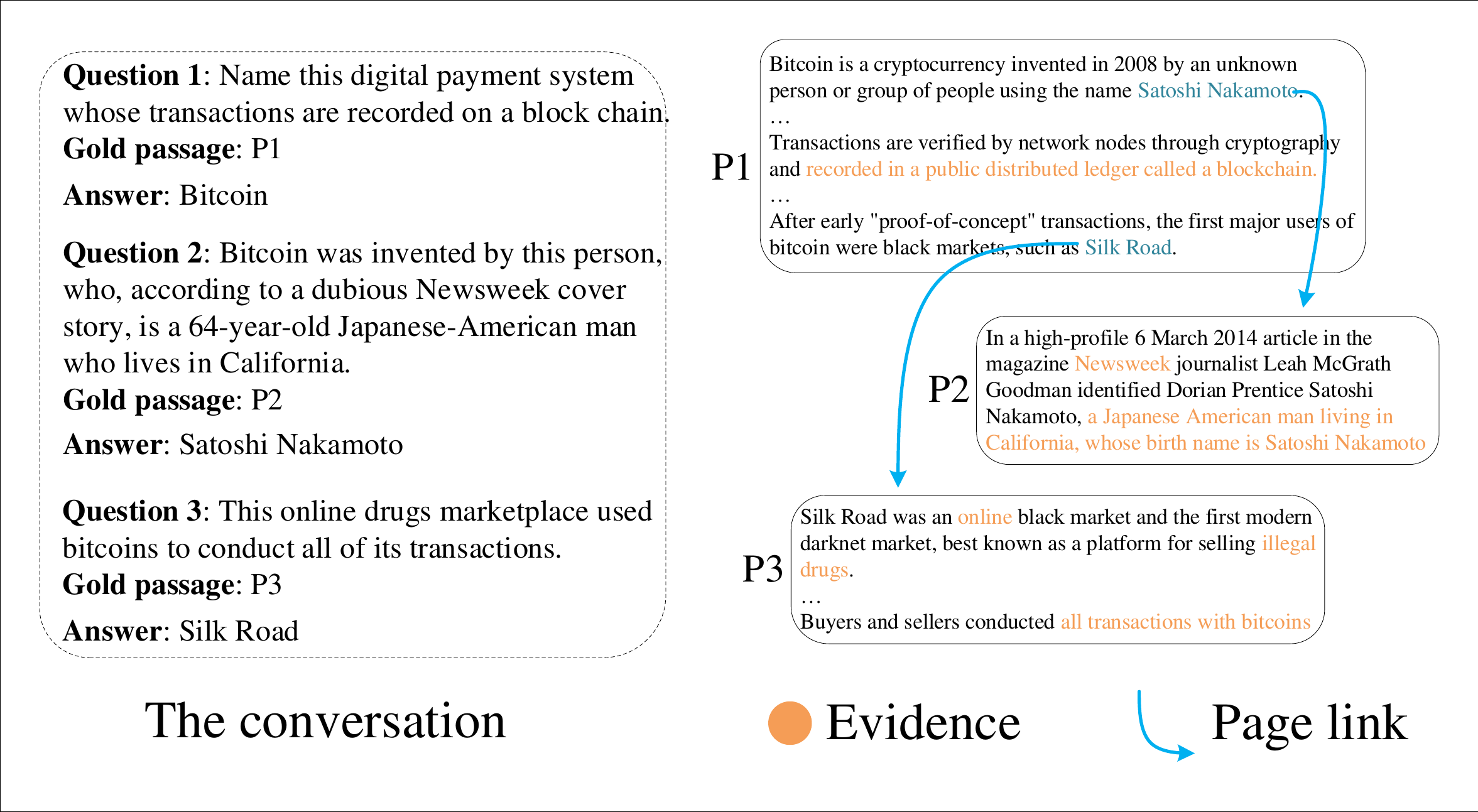}

  \caption{Illustration of the relation among answers in a conversation.}

  \label{example}
\vspace{-1em}
\end{figure}

A typical solution to the problem of conversational open-domain QA is decoupled into historical context modeling, followed by a traditional 3-R open-domain QA pipeline: \textit{Retriever}, \textit{Ranker}, and \textit{Reader}. Historical context modeling plays an important role in understanding the current question with reference to its historical context. With it in place, this problem is then transformed to single-turn QA and handled by the traditional QA techniques~\cite{wang2019document,lu2019answering,zheng2019human}. On historical context modeling, existing conversational open-domain QA models simply pre-append history questions (and sometimes answers too) to the current question~\cite{elgohary-etal-2018-dataset,Qu0CQCI20}. They, however, somehow overlook the relations 
between history answers and the current answer. As shown in Figure~\ref{example}, the gold passages containing answers in a conversation are often highly related and connected via hyperlinks. We also observe that in the QBLink dataset~\cite{elgohary-etal-2018-dataset} about 60\% answers can be found in the two-hop connected passages containing history answers. To address this problem, we introduce a new component called \textit{Explorer} into the existing 3-R pipeline. Underlying the \textit{Explorer} is a novel graph-guided retrieval method. Following \textit{Retriever}, it aims to boost retrieval coverage by capturing the relations among answers across conversation turns via a passage graph. In particular, {\it Explorer} first constructs an initial passage graph with the passages containing history answers and the passages retrieved by {\it Retriever}. It then expands the initial graph by walking $k-$hops via hyperlinks to include more potentially relevant passages. Afterward, a new list of passages is retrieved by applying a Graph Attention Network (GAT) model on the passage graph and the list is passed to the following {\it Ranker} and {\it Reader} to extract the most probable answer.

\begin{figure}

  \includegraphics[width=\linewidth]{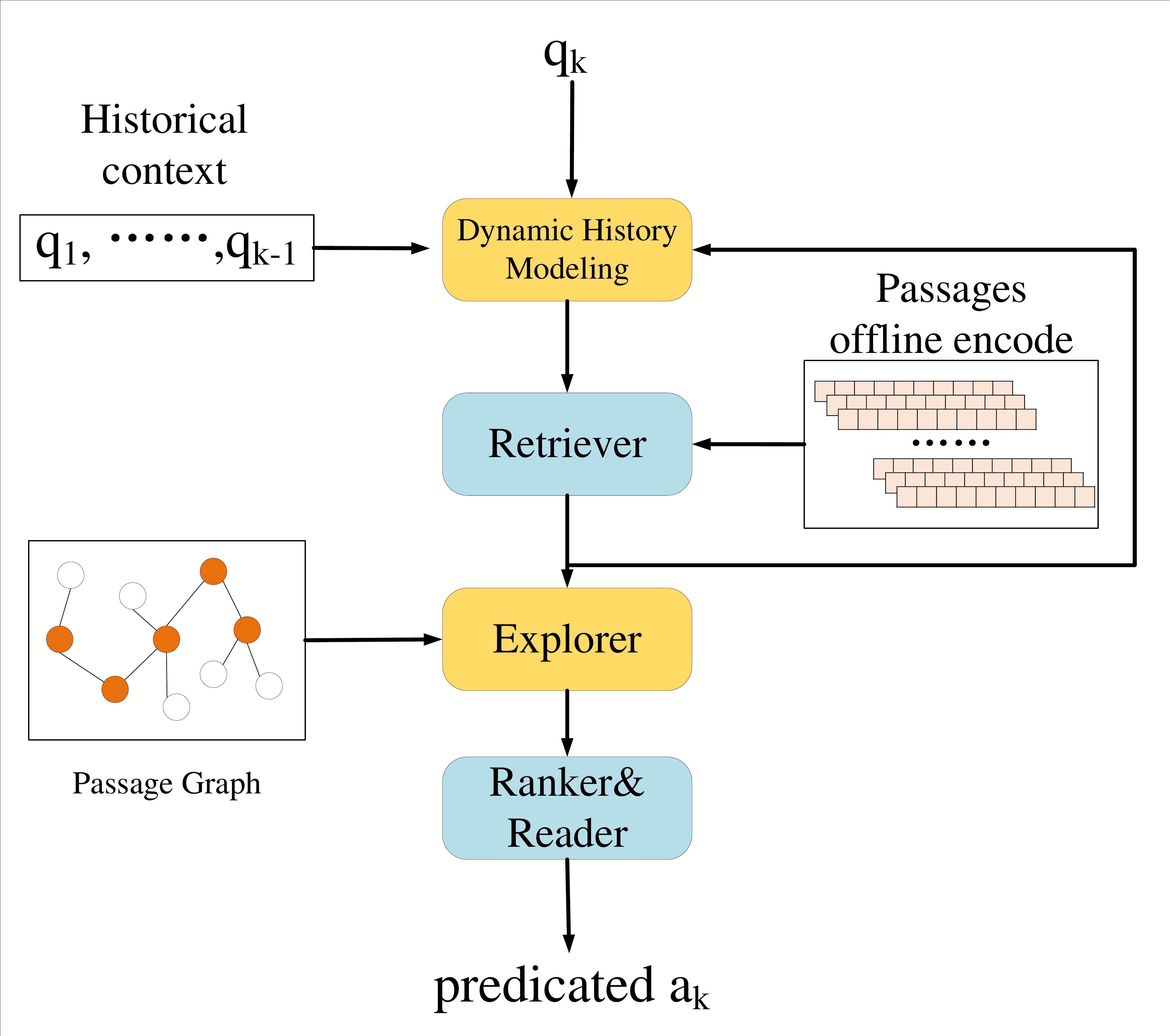}
\vspace{-2em}
  \caption{Illustration of our proposed  pipeline, which contains DHM, Retriever, Explorer, Ranker, and Reader components.}

  \label{framework}
\vspace{-1em}
\end{figure}

As aforementioned, the conversational open-domain QA involves multiple components. Therefore, how to design an effective pipeline, where the components can work collaboratively and interact closely deserves careful consideration. A few existing methods have applied multi-round retrieval methods to open-domain QA, which allow the retriever to receive feedback signals from either the retriever~\cite{feldman2019multi} or the reader\cite{das2018multi}, and generate new queries to update retrieval results. This mechanism can be regarded as relevance feedback~\cite{harman1992relevance}, which has been well explored in traditional information retrieval. However, the impact of relevance feedback on historical context modeling in conversational QA is not yet studied. Inspired by this idea, we propose dynamic history modeling (DHM) to incorporate relevance feedback to enhance the typical historical context modeling component. In particular, DHM considers not only the historical context and the current question but also the candidate passages returned by the retriever, which can be deemed as the retrieval context. The attention mechanism is then adopted to screen out more relevant information in the historical context according to the retrieval context to enhance contextual question understanding. With multi-rounds of retrieval and feedback loops, question embeddings are dynamically refined and improved retrieval performance can be expected.

To summarize, in this paper we propose a novel graph-guided and multi-round retrieval method to enhance the existing conversational open-domain QA pipeline. Specifically, we incorporate a graph-based explorer and a feedback-based DHM component into the pipeline, as illustrated in Figure~\ref{framework}. The key contributions of this work are three-fold:
\begin{itemize}
  \item	We propose a novel graph based \textit{Explorer}, which models relations among answers across conversation turns via the passage graph. By expanding history answers and passages retrieved by the retriever via the hyperlink structure, more relevant passages can be discovered for subsequent answer extraction. 
  \item We incorporate the relevance feedback technique in historical context modeling to iteratively refine current question understanding with reference to the retrieval context. Multi-round retrieval allows us to acquire more relevant passages based on dynamic question embeddings.
  \item Extensive experiments on the public dataset show significant performance improvement over baselines by 5\% and 11\% under two different settings, i.e., with predicted history answers and true history answers, respectively. We will release the data and codes of this work to facilitate other researchers. 
\end{itemize}

\section{Related work}
Our work is closely related to conversational question answering and open-domain question answering.
\subsection{Conversational QA}
Conversational question answering (QA) is an emerging topic in QA. It has been formulated as the tasks of conversational knowledge-based QA (KBQA)~\cite{saha2018complex,christmann2019look,shen2019multi}, conversational search~\cite{dalton2020cast,ren2020conversations}, conversational machine reading comprehension (MCR)~\cite{choi-etal-2018-quac,reddy2019coqa}, and conversational open-domain QA~\cite{elgohary-etal-2018-dataset,Qu0CQCI20}. It is worth emphasizing that the first one relies on knowledge graphs, and the rest three tasks reply questions based upon documents.  Conversational search allows users to interact with the search engine to find the documents that contain potential answers to their questions. Different from conversational search, the gold passage that contains true answers is given in conversational MCR, and it further extracts a text span as the exact answer from the given passage. Comparatively, conversational open-domain QA is more challenging in a sense that it not only retrieves the relevant passages from the Web but also extracts the text answer spans.  

The common problem that all above-mentioned conversational QA tasks have to address is how to understand the historical context in the conversation. For example, there are often questions with ellipsis and/or coreference resolution problems. The common solution in the existing methods is to utilize the historical context to reformulate the current question or refine its representation. A simple strategy is to pre-append all the historical context~\cite{elgohary-etal-2018-dataset} or heuristically select some words or sentences from the context~\cite{choi-etal-2018-quac} to expand the current question. Elgohary et al.~\cite{elgohary2019can} constructed the CANARD dataset, which is able to transform a context-dependent question into a self-contained question. Based on this dataset, a series of question rewriting methods have been proposed. Specifically, some regard question rewriting as a sequence-to-sequence task to incorporate the context into a standalone question. For example, Vakulenko et al.~\cite{vakulenko2020question} used both retrieval and extractive QA tasks to examine the effect of sequence-to-sequence question rewriting on the end-to-end conversational QA performance. Differently, Voskarides et. al.~\cite{VoskaridesLRKR20} modeled query resolution as a binary term classification problem. For each term appearing in the previous turns of conversation, it decides whether or not to add this term to the question in the current turn. Given the gold passage in conversational MCR, the authors in~\cite{qu2019attentive} designed a history attention mechanism to implement ``soft selection'' from conversation histories. The current conversational QA methods have presented various ways to modeling the historical context, and utilizing the historical context to enhance the current question representation. However, they ignore relations among answers (or answer passages) across the conversation turns.  Some efforts~\cite{huang2018flowqa,0022WZ20} in conversational MCR captures the conversational flow by incorporating intermediate representations generated during the process of answering previous questions. Differently, we attempted to model relations more directly and explicitly by exploring the passage graph.
\subsection{Open-domain QA}
Open-domain question answering (QA)~\cite{chen-etal-2017-reading,AAAI1816712,wang2018evidence}, which aims to answer questions from the Web, has attracted wide attention from the academic field in recent years. In 2017, Chen et al.~\cite{chen-etal-2017-reading} first introduced neural methods to open-domain QA using a textual source. They proposed DrQA, a pipeline model with a TF-IDF based retriever and a neural network based reader that was trained to find an answer span given a question and a retrieved paragraph. Later, Wang et al. ~\cite{AAAI1816712} added a ranker between the retriever and reader to  rank the retrieved passages more precisely in 2018. Our proposed method also follows the retriever, ranker and reader pipeline. In addition to TF-IDF based retrievers, dense based retrievers have also been  well developed recently~\cite{lee-etal-2019-latent, karpukhin-etal-2020-dense,chang2019pre}, where all passages are offline encoded in advance to allow efficient large-scale retrieval. In~\cite{karpukhin-etal-2020-dense},  the authors show that the retriever can be practically implemented using dense representations in open-domain QA, where embeddings are learned from a small number
of questions and passages. Our method applies both dense based and TF-IDF based retrievers. The selected top-ranked passages are further extended and explored based on the structure of our constructed  passage graph by our proposed explorer.

There are also researchers exploiting the graph-based retriever for open-domain QA. Min et al. ~\cite{abs-1911-03868} utilized the available knowledge base to build a passage graph and proposed a graph retriever to encode passages. In~\cite{Asai2020Learning}, authors  introduced a graph based recurrent retrieval approach that is learned to retrieve reasoning paths over the Wikipedia graph to answer multi-hop open-domain questions. Distinctly, we aimed to capture the conversation flow with the help of the passage graph. Recently, multi-round retrieval methods have been proposed~\cite{feldman2019multi,das2018multi,xiong2020answering}. The authors in~\cite{feldman2019multi} proposed the multi-hop retrieval method, which  iteratively retrieved supporting paragraphs by forming a joint vector representation of both question and returned paragraph. Das et al.~\cite{das2018multi} proposed a gated recurrent unit to update the question at each round conditioned on the state of the reader, where the retriever and the reader iteratively interact with each other.  Our proposed method also includes a multi-round retrieval mechanism. Beyond existing methods, our focus is to explore relevance feedback, particularly the returned passages, to screen out the most useful information in the historical context to better formulate the current question.

\begin{figure}

  \includegraphics[width=\linewidth]{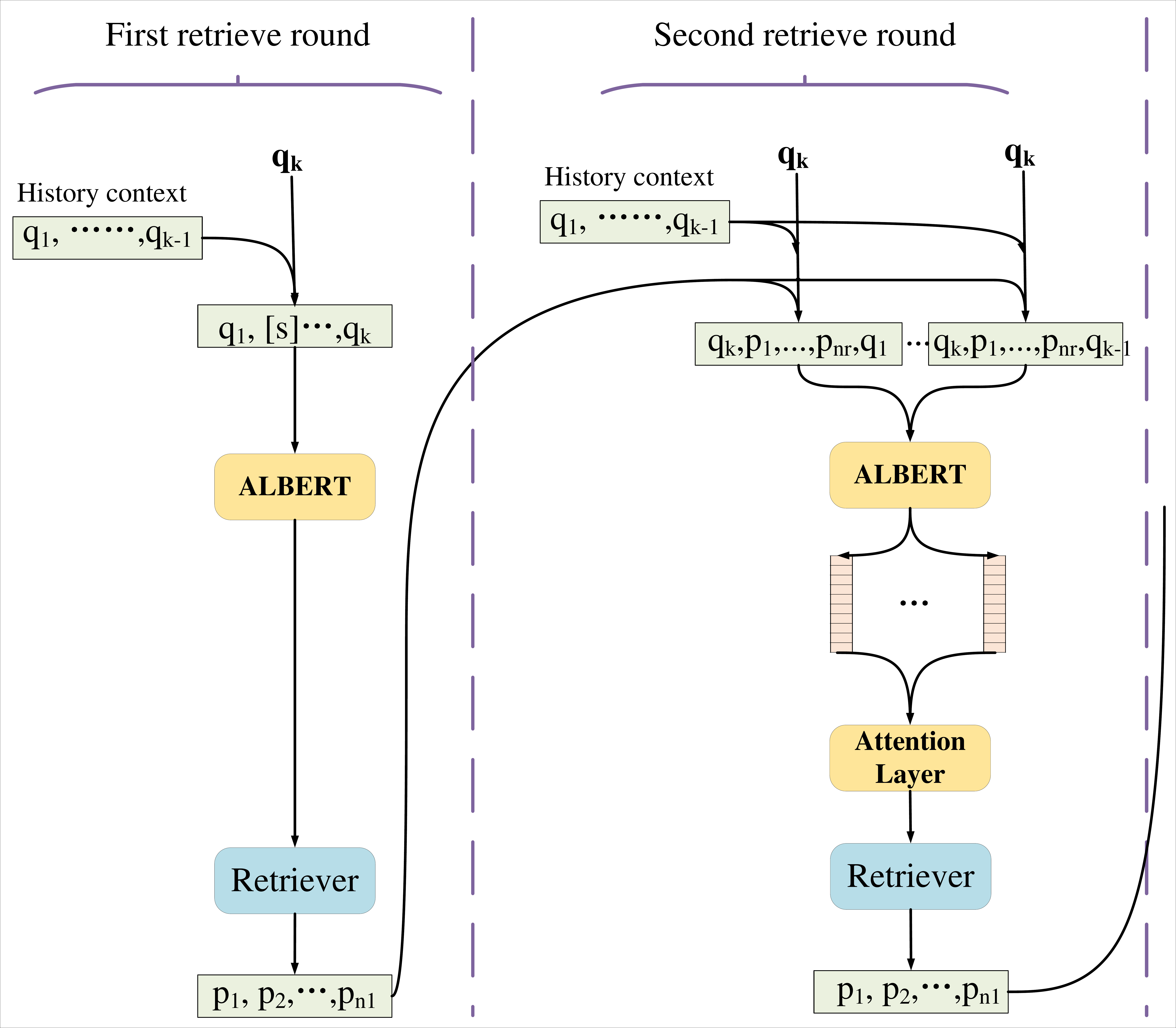}

  \caption{Illustration of the DHM and Retriever. This is a multi-round retrieval method, where  DHM utilizes the relevance feedback to attend useful information in historical context, and the Retriever retrieves a list of passages.}

\vspace{-1.5em}
  \label{DHM}
\end{figure}

\section{Our Proposed Method}
As illustrated in Figure~\ref{framework}, our proposed method comprises the following five components. DHM receives as input the current question $q_k$, the historical context $H_{k}=\{q_1, q_2, …, q_{k-1}\}$\footnote{It is noticed that the history answers are also included in $H_k$ if the true history answers are allowed to use.}, and the passages returned from the retriever if any. It generates a query vector representation via an attention mechanism. The {\it Retriever} retrieves a list of passages $\mathcal{P}_{retriever}$  that is determined relevant to the question representation vector from a large passage collection $\mathcal{C}$. The {\it Explorer} then utilizes the passages received from the {\it Retriever} and  the passages containing history answers to construct an initial passage graph, and expands from the initial passage graph via hyperlink structure to cover more potentially relevant passages. The representations of passages in the passage graph are then updated with a graph neural network model. And a new list of passages $\mathcal{P}_{explorer}$ is retrieved and passed to the subsequent {\it Ranker} and {\it Reader} to re-rank and read the passages to extract the text answer span. We elaborate each component in the following sections.

\subsection{DHM and Retriever}
Our proposed method contains multi-rounds of retrieval, as shown in Figure~\ref{DHM}. In the first round, given the current question $q_k$ and historical context $H_{k}$, DHM encodes the text as a representation vector and then feeds it to the retriever from which the feedback, i.e., retrieved passages $\mathcal{P}_{retriever}$ is received. In the following rounds, DHM utilizes $\mathcal{P}_{retriever}$ and current question $q_k$ to attend to history $H_k$ towards the representation vector refinement iteratively. 

\textbf{The First Round of Retrieval}. We pre-append all history questions $\{q_1, q_2, …, q_{k-1}\}$ to the current question $q_k$, denoted as $q_k^{*}$. To accommodate the BERT~\cite{devlin2019bert} based question encoder, we introduce two special tokens, [CLS] and [SEP]. $q_k^{*}$ is represented as a text sequence ``[CLS] $q_1$ [SEP], $\ldots$, $q_{k-1}$ [SEP] $q_k$ [SEP]''. We use ALBERT~\cite{lan2019albert} to obtain the first-round question representation vector $\textbf{v}_q$ , which is formulated as,
\begin{equation}  \label{eqn1}
   \begin{aligned}
   \textbf{v}_q=\textbf{W}_{q}\textbf{F}_q(q_k^{*}),
   \end{aligned}
 \end{equation}
where $\textbf{F}_q$ is the ALBERT based question encoder, $\textbf{W}_{q}$ is the question projection matrix, and $\textbf{v}_q \in \mathbb{R}^{d_q}$. 
Similarly, for each passage $p_i$ in $\mathcal{C}$ we obtain its representation $\textbf{v}_{p_i}$ as follows,
\begin{equation}  \label{eqn2}
   \begin{aligned}
   \textbf{v}_{p_i}=\textbf{W}_{p}\textbf{F}_p(p_i),
   \end{aligned}
 \end{equation}
where $\textbf{F}_p$ is the ALBERT based passage encoder, $\textbf{W}_{p}$ is the passage projection matrix, $\textbf{v}_{p_i} \in \mathbb{R}^{d_p}$, and $d_p$ is equal to $d_q$. It is critical that paragraph encodings are independent of questions in order to enable storing precomputed paragraph encodings and executing the efficient maximum inner product search (MIPS) algorithm~\cite{lee-etal-2019-latent, karpukhin-etal-2020-dense}. Otherwise, any new question would require re-processing the entire passage collection (or at least a significant part of it). Benefiting from this, we can calculate the similarity score via the inner product of $\textbf{v}_q$ and each passage $\textbf{v}_{p_i}$ efficiently, and select the top-$n_1$ passages $\mathcal{P }_{retriever}=\{p_1$, $p_2$, $\ldots$, $p_{n_{1}}\}$, where $n_1$ is the number of the passages in $\mathcal{P }_{retriever}$.

\textbf{The Following Rounds of Retrieval}. The returned passages in $\mathcal{P}_{retriever}$  are incorporated into the history questions $H_k$ and the current question $q_k$ in the form of triplet $\{q_i^t\}_{i=1}^{k-1}$, where each triplet $q_i^t$ contains the current question $q_k$, the top-$n_r$ passages in $\mathcal{P }_{retriever}$, and the $i$-th history question. It is formulated as $q_i$ ``[CLS] $q_k$ [SEP] $p_1$, \ldots, [SEP] $p_{n_r}$ [SEP] $q_i$ [SEP]''. As such, interactions among the history question, the current question and the retrieval feedback are encouraged. The representation of $q_i^t$ is calculated as follow,
\begin{equation}  \label{eqn3}
   \begin{aligned}
   \textbf{v}_{q_i^t}=\textbf{W}_{q}\textbf{F}_q(q_i^t).
   \end{aligned}
 \end{equation}
A history attention network is then followed to aggregate the $k-1$ representation vectors into the refined question vector $\textbf{v}_{q}$ with learned attention weights. Formally, the attention layer is defined as follows,
\begin{equation}\label{eqn4}
  \left\{
   \begin{aligned}
& \alpha_{i}  =\frac{\exp(\textbf{W}_a {\textbf{v}_{q_i^t})}}
{\sum_{i=1}^{k-1}\exp(\textbf{W}_a {\textbf{v}_{q_i^t})} }, \\
& \textbf{v}_{q} = \sum_{i=1} ^{k-1}{ \alpha_{i}} \textbf{v}_{q_i^t},
   \end{aligned}
   \right.
  \end{equation}
where $\textbf{W}_a \in \mathbb{R}^{1 \times d_q}$, $\alpha_{i}$ denotes
the attention weight of the $i$-th triplet, and $\textbf{v}_q \in \mathbb{R}^{d_q}$. The attention weights help to capture certain contexts that are more relevant to the current round of retrieval. With the new question representation $\textbf{v}_{q}$, we re-calculate the similarity scores and select the new list of top-$n_1$ passages as before. The question representation vector $\textbf{v}_q$ is continuously updated when the returned passage changes in each round. After several rounds, the final rank list of passages $\mathcal{P}_{retriever}$ is delivered to the following Explorer.

\subsection{Explorer}
\begin{figure}
  \includegraphics[width=\linewidth]{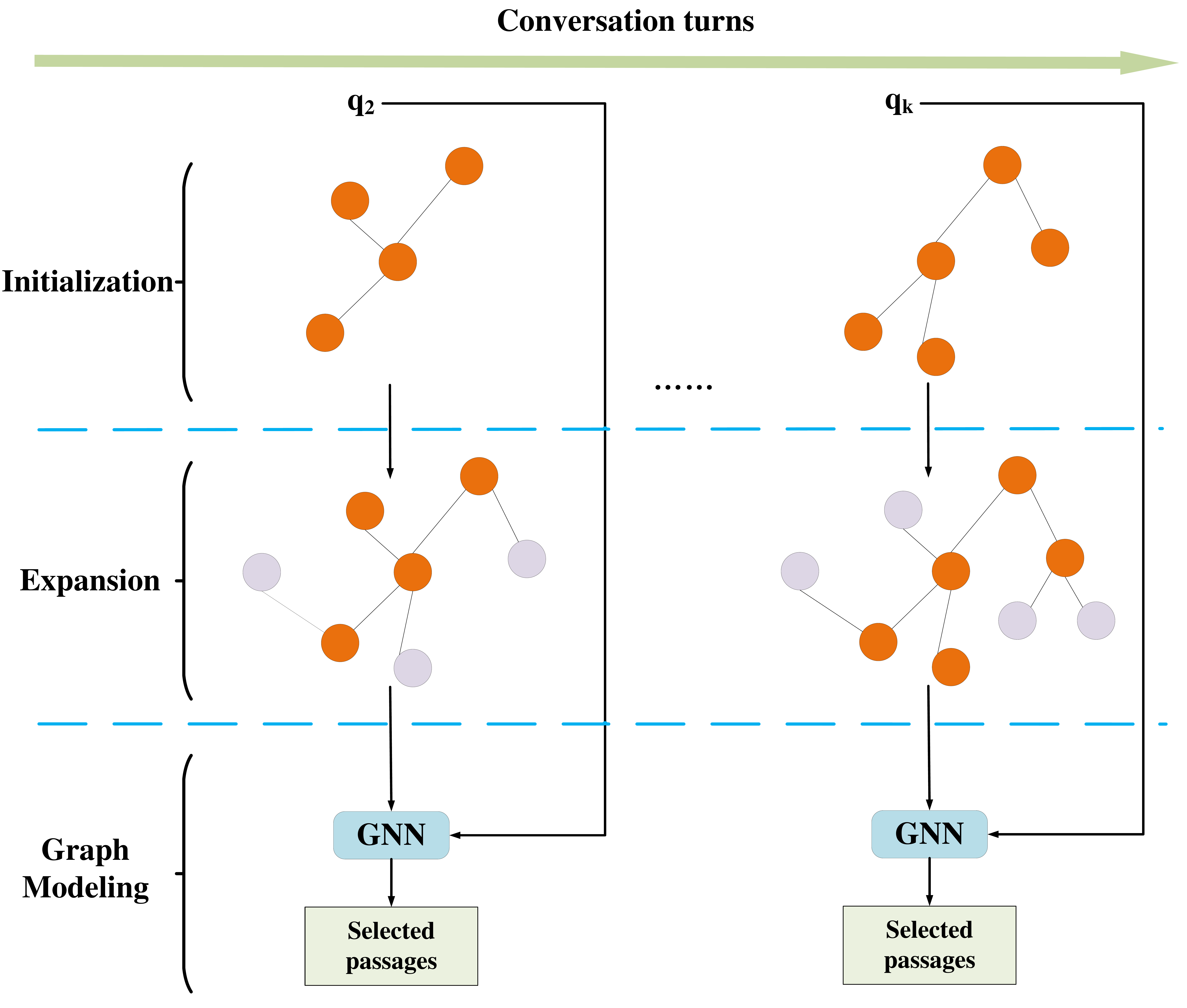}

  \caption{Illustration of Explorer component.}
  \label{Explorer}
\end{figure}

Essentially, our proposed explorer is equipped with a graph-guided candidate answer passage expansion method. Based on the observation that answers are also highly related in conversation QA, we design {\it Explorer} to model relations of answers by constructing a passage graph. In general, the explorer goes through the following three steps: initialization, expansion, and graph modeling, as illustrated in Figure~\ref{Explorer}.

We first construct an initial passage graph $\mathcal{G}_0=<\mathcal{P}_0, \mathcal{E}_0>$, where $\mathcal{P}_0$ are some selected passages and $\mathcal{E}_0$ are hyperlinks among them. $\mathcal{P}_0$ includes the passages containing gold/predicted answers in previous turns of the conversation and the passages retrieved by the dense retriever described in the previous section. To increase the pool of relevant passages, we also run a simple TF-IDF based retrieval model. Altogether we have
\begin{equation}  \label{eqn5}
   \begin{aligned}
   \mathcal{P}_0=\{\mathcal{P}_a \cup \mathcal{P}_{retriever} \cup  \mathcal{P}_T\},
   \end{aligned}
 \end{equation}
where $\mathcal{P}_a$, $\mathcal{P}_{retriever}$, and $\mathcal{P}_T$ denote the set of passages that contain previous gold/predicted answers, the passages returned by the dense retriever, and the passage returned by the TF-IDF retriever, respectively. Starting from the initial passage nodes $\mathcal{P}_0$, we expand the passage set to $\mathcal{P}_m$  by adding the hyperlink-connected passages via,
\begin{equation}\label{eqn6}
   \begin{aligned}
\mathcal{P}_m=\{\mathcal{P}_{m-1} \cup \mathcal{P}_e\}, 
   \end{aligned}
  \end{equation}
where $ \mathcal{P}_e$ denotes the passages linked to the passages in $\mathcal{P}_{m-1}$ via the Web hyperlinks. Through expansion, a larger set of passages $\mathcal{P}_m=\{p_i\}_{i=1}^{n_m}$ is obtained, where $n_m$ refers to the number of passages in $\mathcal{P}_m$. Meanwhile, we obtain the corresponding expanded passage graph $\mathcal{G}_m$, whose nodes are passages in $\mathcal{P}_m$ and edges are hyperlinks among them.

To take advantage of the structure information, we apply a Graph Attention network~\cite{velivckovic2018graph} to update the embeddings of the passages in $\mathcal{G}_m$. For each passage $p_i$ in $\mathcal{G}_m$, we input it to an ALBERT encoder to obtain its  embedding by Eqn.($\ref{eqn2}$) and then update the node embedding as follows,
\begin{equation}  \label{eqn7}
   \begin{aligned}
   {\textbf{v}_{p_i}}^*=\textbf{GAT}(\textbf{v}_{p_i},\mathcal{G}_m).
   \end{aligned}
 \end{equation}
Based on the updated embeddings, we re-calculate the similarity between the question and the passages in $\mathcal{G}_m$. Specifically, we obtain the question embedding $\textbf{v}_q$ by Eqn.($\ref{eqn1}$) and calculate the similarity $S_{a_i}$ of passage $p_i$  to $\textbf{v}_q$ as follows,

\begin{equation}  \label{eqn8}
   \begin{aligned}
   S_{a_i} =\frac{\exp({\textbf{v}_{p_i}}^*\textbf{v}_q^T)}
{\sum_{i=1}^{n_m}\exp({\textbf{v}_{p_i}}^*\textbf{v}_q^T)}.
   \end{aligned}
 \end{equation}
Finally, the top-$n_2$ passages $\mathcal{P}_{explorer}$ are selected according to their ranks based on similarity calculation, where $n_2$ denotes the number of passages in $\mathcal{P}_{explorer}$. $\mathcal{P}_{explorer}$ is then passed to the following {\it Ranker} and {\it Reader} for further processing.

\subsection{Ranker and Reader}
Given the list of passages $\mathcal{P}_{explorer}$, {\it Ranker} re-ranks the passages more precisely and {\it Reader} predicts a text span as an answer. We first construct the reformatted text sequence $[q, p_i]$ by concatenating the question text and the passage $p_i$ text. $[q, p_i]$ is ``[CLS] $q_k^*$ [SEP] $p_i$ [SEP]''. For each token $token_j$ in $[q, p_i]$, a BERT encoder generates its representation $\textbf{v}_{t_j}$, and for the whole text sequence, BERT encoder generates its representation $\textbf{v}_{q,p_i}$.

{\it Ranker} conducts a listwise reranking of the top-$n_2$ passages received from {\it Explorer} and calculates the reranking score for each passage $p_i$ in $\mathcal{P}_{explorer}$ as,

\begin{equation}  \label{eqn9}
   \begin{aligned}
   S_{b_i} =\frac{\exp(\textbf{v}_{q,p_i}\textbf{W}_{ra})}
{\sum_{i=1}^{n_m}\exp(\textbf{v}_{q,p_i}\textbf{W}_{ra})},
   \end{aligned}
 \end{equation}
where $\textbf{W}_{ra} \in \mathbb{R}^{d_q \times 1}$. {\it Reader} then predicts an answer span by computing two scores for each token $j$ in passages in $\mathcal{P}_{explorer}$ as the start token and the end token, respectively, formulated as,

\begin{equation}\label{eqn10}
  \left\{
   \begin{aligned}
& S_{s_j}=\frac{\exp(\textbf{v}_{t_j}\textbf{W}_s)}
{\sum_{i=1}^{n_t}\exp(\textbf{v}_{t_j}\textbf{W}_s)},\\
& S_{e_j}=\frac{\exp(\textbf{v}_{t_j}\textbf{W}_e)}
{\sum_{i=1}^{n_t}\exp(\textbf{v}_{t_j}\textbf{W}_e)},
   \end{aligned}
   \right.
  \end{equation}
where $n_t$ is the number of tokens, and both $\textbf{W}_s$ and $\textbf{W}_e \in \mathbb{R}^{d_q \times 1}$. Finally, the text span with the highest score is extracted as the answer. We will detail the inference process in the following section.

\subsection{Training and Inference}
\textbf{Training}. Recall that we encode the large collection of passages $\mathbb{C}$ offline for efficient retrieval. Specifically, we follow the previous work~\cite{lee-etal-2019-latent,Qu0CQCI20} to pretrain a passage encoder so that it can provide reasonably good retrieval results to the subsequent components for further processing. After offline encoding, a set of passage vectors are obtained. Note that while the parameters of the passage encoder $\textbf{F}_p$ are fixed after pretraining, the parameters of the question encoder $\textbf{F}_q$ are updated along with those of the explorer, the ranker and the reader.

After passage offline encoding, we then train the retriever, the explorer, the ranker, and the reader. For each passage $p_i$ in  $\mathcal{P}_{retriever}$, we calculate the retrieval score $S_{re_i}$ as follows,

\begin{equation}  \label{eqn11}
   \begin{aligned}
   S_{re_i} =\frac{\exp(\textbf{v}_{q}{\textbf{v}_{p_i}}^T)} 
{\sum_{i=1}^{n_1}\exp(\textbf{v}_{q}{\textbf{v}_{p_i}}^T)}.
   \end{aligned}
 \end{equation}
To train the retriever, we set the retriever loss $L_{retriever}$ as follows,
\begin{equation}  \label{eqn12}
   \begin{aligned}
   L_{retriever}=-\sum_{i=1}^{n_1}(y \log(S_{re_i})+(1-y) \log(1-(S_{re_i}))),
   \end{aligned}
 \end{equation}
where $y$ indicates whether the passage is a gold passage. If the gold passage is not presented in $\mathcal{P}_{retriever}$, we manually include it in the retrieval  results.
Similar to the retriever loss, we respectively define the explorer loss $L_{explorer}$ and the ranker loss $L_{ranker}$ as follows,

 \begin{equation}\label{eqn13}
  \left\{
   \begin{aligned}
& L_{explorer}=-\sum_{i=1}^{n_m}(y \log(S_{a_i})+(1-y) \log(1-(S_{a_i}))),\\
& L_{ranker}=-\sum_{i=1}^{n_m}(y \log(S_{b_i})+(1-y) \log(1-(S_{b_i}))),
   \end{aligned}
   \right.
  \end{equation}
  
 where $S_{a_i}$ and $S_{b_i}$ are the explorer score and ranker score of the passage $p_i$ in $\mathcal{P}_{explorer}$, respectively. The reader loss $L_{reader}$ is formulated as,
\begin{equation}  \label{eqn14}
   \begin{aligned}
   L_{reader}=-\sum_{j=1}^{n_t}(y_1 \log(S_{s_j})+(1-y_1) \log(1-(S_{s_j})))\\-\sum_{j=1}^{n_t}(y_2 \log(S_{e_j})+(1-y_2) \log(1-(S_{e_j}))),
   \end{aligned}
 \end{equation} 
  where $y_1$ and $y_2$ indicate whether the token is the start token and the end token, respectively. Considering the limitation of GPU memory, we first train the retriever, the ranker, and the reader jointly via the sum of their losses, and then train DHM and the explorer separately.

\textbf{Inference}. For each passage in $\mathcal{P}_{explorer}$, we obtain the explorer score $S_a$ and the ranker score $S_b$ by Eqn.($\ref{eqn8}$) and Eqn.($\ref{eqn9}$), respectively. For each token, the reader then assigns it the probabilities of being the start token $S_s$ and the end token $S_e$. Following the convention~\cite{devlin2019bert,Qu0CQCI20}, we consider the top 20 text spans only to ensure tractability. Invalid predictions, including the cases where the start token comes after the end token, or the predicted span overlaps with a question in the conversation context, are all discarded. The predicted score $S$ of a potential answer is then calculated as,
 \begin{equation}  \label{eqn15}
   \begin{aligned}
   S=S_a+S_b+S_s+S_e.
   \end{aligned}
 \end{equation} 
The model finally outputs the answer span with the maximum overall score to respond to the current question.

\begin{table}[tbp]

\centering
\caption{Statistics of the dataset.}\label{dataset statistics}
\vspace{-0.5em}
 \scalebox{0.9}{
\begin{tabular}{ccccc}
\toprule
&items&Train&Dev&Test\\
\midrule
\multirow{3}{*}{QA pairs} &\# Dialogs& 4,383&490&771\\
&\# Questions&31,526&3,430&5,571\\
&\# Avg.Questions
per Dialog &7.2& 7.0& 7.2\\
\midrule
Collection&\# passages &\multicolumn{3}{c}{11 million}\\
\midrule
\multirow{3}{*}{Passage Graph}&\# nodes &\multicolumn{3}{c}{11 million}\\
&\# edges &\multicolumn{3}{c}{105 million}\\
&\# Avg. edges per node &\multicolumn{3}{c}{17}\\
\bottomrule
\end{tabular}}
\vspace{-1em}
\end{table}

\section{Experiments}
\subsection{Dataset}
To conduct experiments, we used the public available dataset OR-QuAC ~\cite{Qu0CQCI20}, which expands the QuAC dataset~\cite{choi-etal-2018-quac} to the open-domain QA setting. Each conversation in this dataset contains a series of questions and answers, and is supplemented with a collection of passages. There are totally 5,644 conversations covering 40,527 questions. The passage collection is from the English Wikipedia dump file of 10/20/2019, and it contains about 11 million passages. We further crawled the internal page links in the Wikipedia via Wikiextractor\footnote{https://github.com/attardi/wikiextractor.} to facilitate passage graph construction. The statistics of the above dataset are summarized in Table~\ref{dataset statistics}. 
\subsection{Experimental Settings}
\textbf{Evaluation Protocols}. Following the previous work~\cite{choi-etal-2018-quac,Qu0CQCI20}, we employed the word-level F1 and the human equivalence score (HEQ), which are
two metrics provided by the QuAC challenge~\cite{choi-etal-2018-quac} to evaluate the Conversational QA systems. HEQ computes the percentage of examples for which system F1 exceeds or matches human F1. It measures whether a system can
give answers as good as an average human. This metric is computed
on a question level (HEQ-Q) and a dialog level (HEQ-D). Besides, to evaluate the retrieval performance of the retrieval passage list, we also applied the Mean Reciprocal Rank
(MRR) and Recall to evaluate the Retriever component of the baselines, our Explorer component, and the Ranker component. 

\textbf{Implementation Details}. We utilized the pretrained passages embedding vectors released in~\cite{Qu0CQCI20} to make a fair comparison. The $d_q$, $d_p$ are both set to 128, and $n_2$ is set to 5, the same as~\cite{Qu0CQCI20}. The number of retrieved passages from the retriever $n_1$ is 3. Limited by our GPU memory, both $m$ and $n_r$ is set to 1, and the retrieve rounds are 2. There are two GAT layers, where the numbers of heads are 128 and 64, respectively. The passage embeddings are stored in a 2080Ti card and the model is in another 2080Ti card.
\subsection{Baselines}
To the best of our knowledge, only one published work~\cite{Qu0CQCI20}, ORConvQA, aims to solve the conversational open-domain QA problem. We compared our proposed method with ORConvQA and the baselines in~\cite{Qu0CQCI20} as follows:
\begin{itemize}
\item \textbf{DrQA}~\cite{chen-etal-2017-reading}: This model uses a TF-IDF retriever and a RNN based
reader. The original distantly supervised setting is not applied, since the full supervision is allowed in the dataset and adopted for all the methods.
\item \textbf{BERTserini}~\cite{yang2019end}: This model uses a BM25 retriever implemented in Anserini~\footnote{http://anserini.io/.} and a BERT reader. Their BERT reader is similar to ours, except that it does not support reranking and thus cannot benefit from joint learning. 
\item \textbf{ORConvQA}~\cite{Qu0CQCI20}: It is an open-retrieval conversational question answering method. It uses a dense retriever, ranker, and reader pipeline, which is similar to ours. ORConvQA utilizes a sliding window to append previous questions. 
\item \textbf{ORConvQA without history} (ORConvQA w/o hist.): This is ORConvQA
 with the history window size $w = 0$. However, the first question is always included in the reformulated current question.
\end{itemize}

 Following the previous work~\cite{Qu0CQCI20}, we evaluated all methods under the setting of no true history answers. The results of these baselines are public in~\cite{Qu0CQCI20}. We also evaluated ORConvQA and ours with true history answers, denoted as ORConvQA-ta and ``Ours-ta'', respectively.

\begin{table*}[t]

  \centering
  \caption{Performance comparison between our proposed model and state-of-the-art baselines over develop and test set. ``Rt-R'', ``Rt-M'', and ``Rr-M'' refers to recall of the retriever, MRR of the retriever, and MRR of the ranker, respectively.  It is worth mentioning that we evaluated our Explorer as the retriever.
 * means statistically significant improvement
over the strongest baseline with p < 0.05. }\label{overall_performance}
\vspace{-0.5em}
  \label{tab:Overall Comparison}

    \scalebox{1.0}{
    \begin{tabular}{ccccccccccccc}
    \hline
    \multirow{2}{*}{Methods}&
    \multicolumn{6}{c|}{Dev}&\multicolumn{6}{c}{ Test}\cr\cline{2-13}
         &Rt-R&Rt-M&Rr-M&H-Q&H-D&F1 &Rt-R&Rt-M&Rr-M&H-Q&H-D&F1\cr \hline
    
    DrQA&0.2000&0.1151&N/A&0.0&0.0&4.5&0.2253&0.1574&N/A&0.1&0.0&6.3\cr\hline

    BERTserini&0.2656&0.1767&N/A&14.1&0.2&19.3&0.2507&0.1784&N/A&20.4&0.1&26.0\cr\hline
    
    ORConvQA w/o hist&0.5271&0.4012&0.4472&15.2&0.2&24.0&0.2859&0.1979&0.2702&20.7&0.4&26.3\cr\hline
    
    ORConvQA&0.5714&0.4286&0.5209&17.5&0.2&26.9&0.3141&0.2246&0.3127&24.1&0.6&29.4\cr\hline
    
   \makecell[c]{ORConvQA-ta} &0.6833&0.5414& 0.6392&19.6&0.2&31.4&0.4721& 0.3810& 0.4765&29.6&1.2&34.6\cr\toprule
   
    Ours&0.6338&0.4112&0.5410&17.6&0.2&28.1& 0.3674&0.2043&0.3508&30.3&1.0&33.4\cr\hline
    
   \makecell[c]{Ours-ta} &$\textbf{0.8776}^*$&$\textbf{0.8015}^*$& $\textbf{0.8200}^*$& $\textbf{20.5}^*$&$\textbf{0.4}^*$& $\textbf{32.9}^*$&  $\textbf{0.8395}^*$&$\textbf{0.7854}^*$&$\textbf{0.8336}^*$&$\textbf{39.2}^*$&$\textbf{1.2}$&$\textbf{45.3}^*$\cr\hline
    \end{tabular}}
\vspace{-1em}
\end{table*}

\begin{figure*}[htbp]

  \centering

  \subfigure[ Recall on test set]{

  \includegraphics[width=0.32\linewidth]{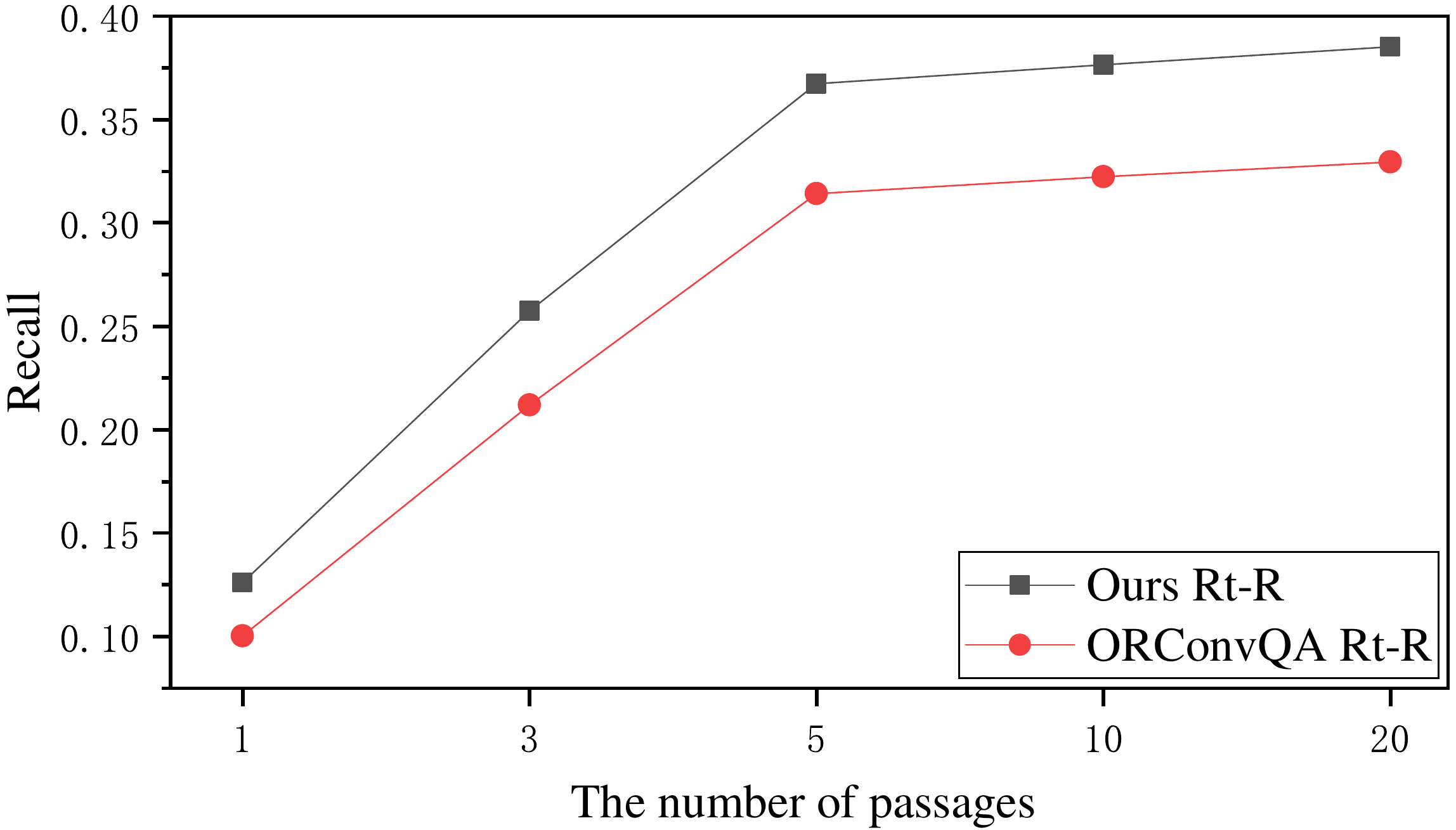}
   \label{fig:subfig1}
   }
   \subfigure[ MRR on test set]{

  \includegraphics[width=0.32\linewidth]{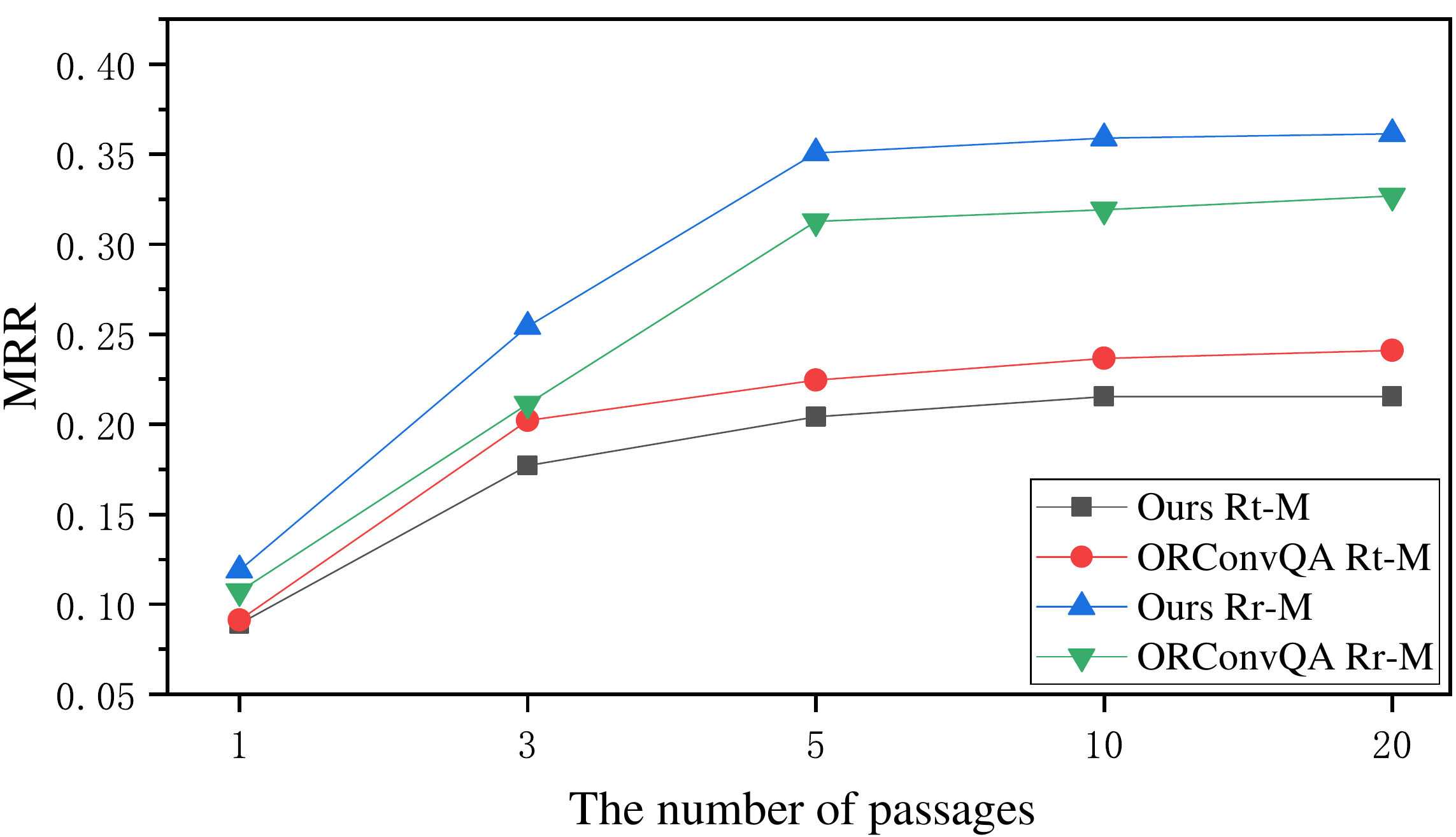}
   \label{fig:subfig2}
   }
   \subfigure[F1 on test set]{

  \includegraphics[width=0.32\linewidth]{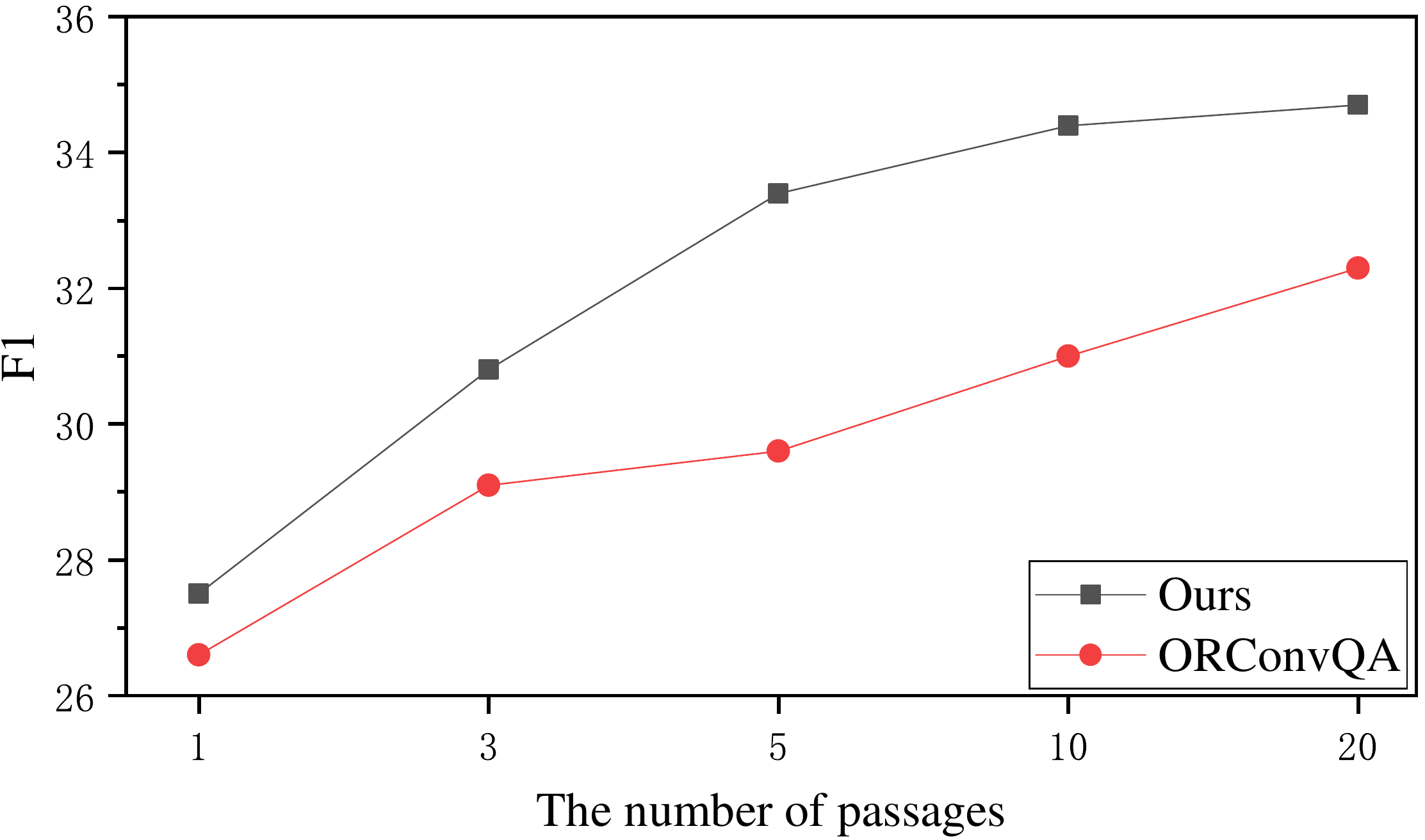}
   \label{fig:subfig3}
   }

\vspace{-1em}
  \caption{Retrieval performance versus the number of returned passages from the Explorer.}\label{overall_fig}

\vspace{-1em}
\end{figure*}

\subsection{Overall Comparison}
The results of all methods are summarized in Table~\ref{overall_performance}, from which we have the following findings.

(1) As expected, traditional single-turn open-domain QA methods, i.e. DrQA and BERTserini, perform worse than the conversational QA methods, since they consider no historical context. And thus they fail to solve the elliptical and coherence problems in conversations. Moever, the methods based on a BERT reader, i.e. BERTserini, ORConvQA, and ``Ours'', have a significant improvement over DrQA, a RNN based reader. This indicates pretrained BERT is more powerful than RNN in producing semantic representations.

(2) The dense retriever based methods, including ORConvQA and ``Ours'', surpass the BM25 retriever based methods. This is also verified in the single-turn QA~\cite{lee-etal-2019-latent, karpukhin-etal-2020-dense}. On the one hand, the semantic similarities can be learned and reflected via the distances in the embedding space. On the other hand, the dense retrievers allow joint learning of different components in the conversational open-domain QA pipeline.

(3) Our method achieves the best performance, substantially surpassing all baselines. In specific, our method has a dense retriever and a BERT reader, outperforming DrQA and BERTserini consistently. Although our method has the same ranker and reader as ORConvQA, our method still considerably surpasses it. This verifies that our proposed graph-guided and multi-round dynamic retrieval method is able to provide a better list of passages containing more relevant passages to the following ranker and reader. 

(4) Both ORConvQA and ``Ours'' with true history answers surpass with predicted history answers. This demonstrates that the history answers indeed provide useful and necessary information for the current question understanding, which is consistent with the observation on the QuAC dataset~\cite{choi-etal-2018-quac}. ``Ours-ta'' significantly outperforms ORConvQA-ta, and it improves the recall to 0.84. That is because that our proposed Explorer fully takes advantages of the history answers by activates them in the passage graph. Therefore, when true history answers available, gold passages can be easily found via the graph structure.  

To justify the performance of the retrieval passage list, we compared the performance of our model and the best baseline ORConvQA by varying the number of retrieved passages $n_2$. From Figure 5, our observations are as follows,

(1) Looking at both Figure 5(a) and 5(b), we can see that when the number of returned passages increases, both Recall and MRR rise. This is because more and more gold passages are retrieved with the increased number of returned passages. We can also find that the F1 score increases fast at the beginning and then gradually degrades. This shows that although increasing the number of returned passages benefits Recall and MRR, it also becomes a heavy burden to the reader as more noisy passages are also delivered to it.

(2) In terms of Rt-R, ``Ours'' outperforms ORConvQA significantly. The retrieved passages of ours contain more gold passages, because our proposed explorer find more relevant passages via the graph structure.  But in terms of Rt-M, ORConvQA exceeds ``Ours''. It may be because of the GAT layer, which allows the embeddings become more similar to the surroundings. Thus, it is easier for our explorer to retrieve more gold passages and their neighbors but cannot precisely distinguish them. However, compared to MRR, improving recall is more important for the retriever in the pipeline, because the following ranker can further re-rank the passages in the list. As shown in Table~\ref{overall_performance}, the Rr-M of ours outperforms that of ORConvQA, demonstrating  that our ranker works well and provides a complementation to our explorer.

\subsection{ Component-wise Evaluation }
We further tested the variants of our model to further verify the effectiveness of the components in the pipeline.
\begin{itemize}
\item \textbf{Ours w/o Ranker}: The ranker is excluded from the pipeline.  And the ranker’s score $S_b$ is removed from the final predicted score in Eqn.($\ref{eqn15}$).
\item \textbf{Ours w/o DHM}: DHM is excluded from the pipeline. That is, the question representation vector is generated as described in the first round of retrieval. Since there is no dynamic history modeling, there is no multi-round retrieval process and no relevance feedback from the retriever.
\item \textbf{Ours w/o Explorer}: The Explorer is excluded from the pipeline. That is, the passage list $\mathcal{P}_{retriever}$ rather than the list $\mathcal{P}_{explorer}$ is delivered to the ranker and the reader. 
\end{itemize}

\begin{table*}[t]

  \centering
  \caption{Component-wise validation of our proposed method by disabling one component each time.}\label{component_wise}
  \vspace{-0.5em}
  \label{tab:Overall Comparison}

    \scalebox{1.1}{
    \begin{tabular}{ccccccccccccc}
\hline
    \multirow{2}{*}{Methods}&
    \multicolumn{6}{c|}{Dev}&\multicolumn{6}{c}{ Test}\cr \cline{2-13}
         &Rt-R&Rt-M&Rr-M&H-Q&H-D&F1 &Rt-R&Rt-M&Rr-M&H-Q&H-D&F1\cr \hline

    Ours w/o DHM&0.5997&0.3988&0.5025&17.6&0.2&27.5&0.3350&0.1661&0.3157&30.3&1.0&32.6\cr\hline
    
    Ours w/o Explorer&0.5846&0.3900&0.4981&17.5&0.0&27.3&0.2494&0.1351&0.2295&27.5&0.9&29.7\cr\hline
    
    Ours w/o Ranker&0.6220&0.4057&N/A&17.6&0.2&27.9&0.3423&0.1892&N/A&20.4&1.0&33.1\cr\hline
   
    Ours&\textbf{0.6338}&\textbf{0.4112}&\textbf{0.5410}&\textbf{17.6}&\textbf{0.2}&$\textbf{28.1}$& \textbf{0.3674}&\textbf{0.2043}&\textbf{0.3508}&\textbf{30.3}&\textbf{1.0}&$\textbf{33.4}$\cr
\hline

    \end{tabular}}

\end{table*}

\begin{figure*}[htbp]

  \centering

  \subfigure[]{

  \includegraphics[width=0.32\linewidth]{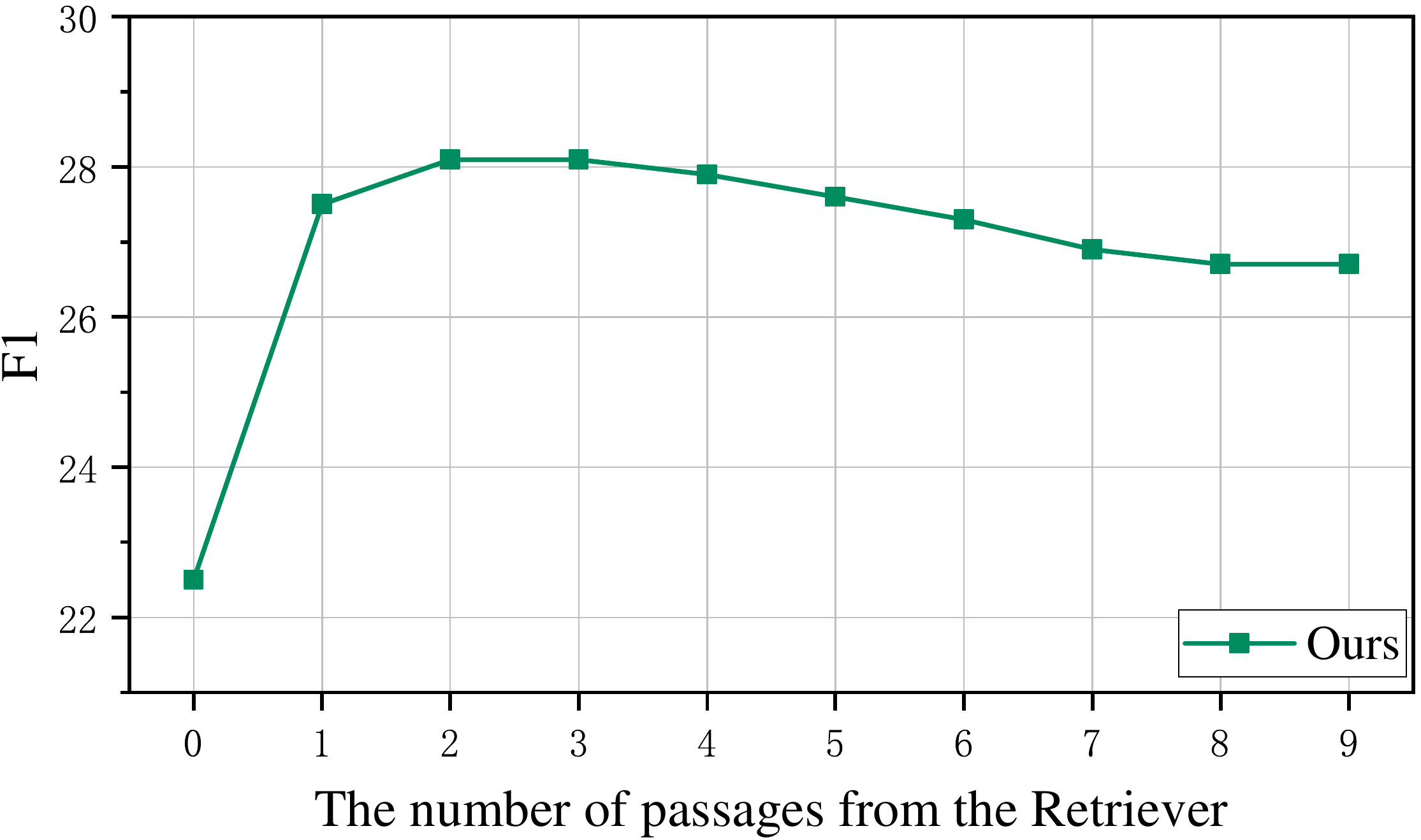}
   \label{fig6:subfig1}
   }
   \subfigure[]{

  \includegraphics[width=0.32\linewidth]{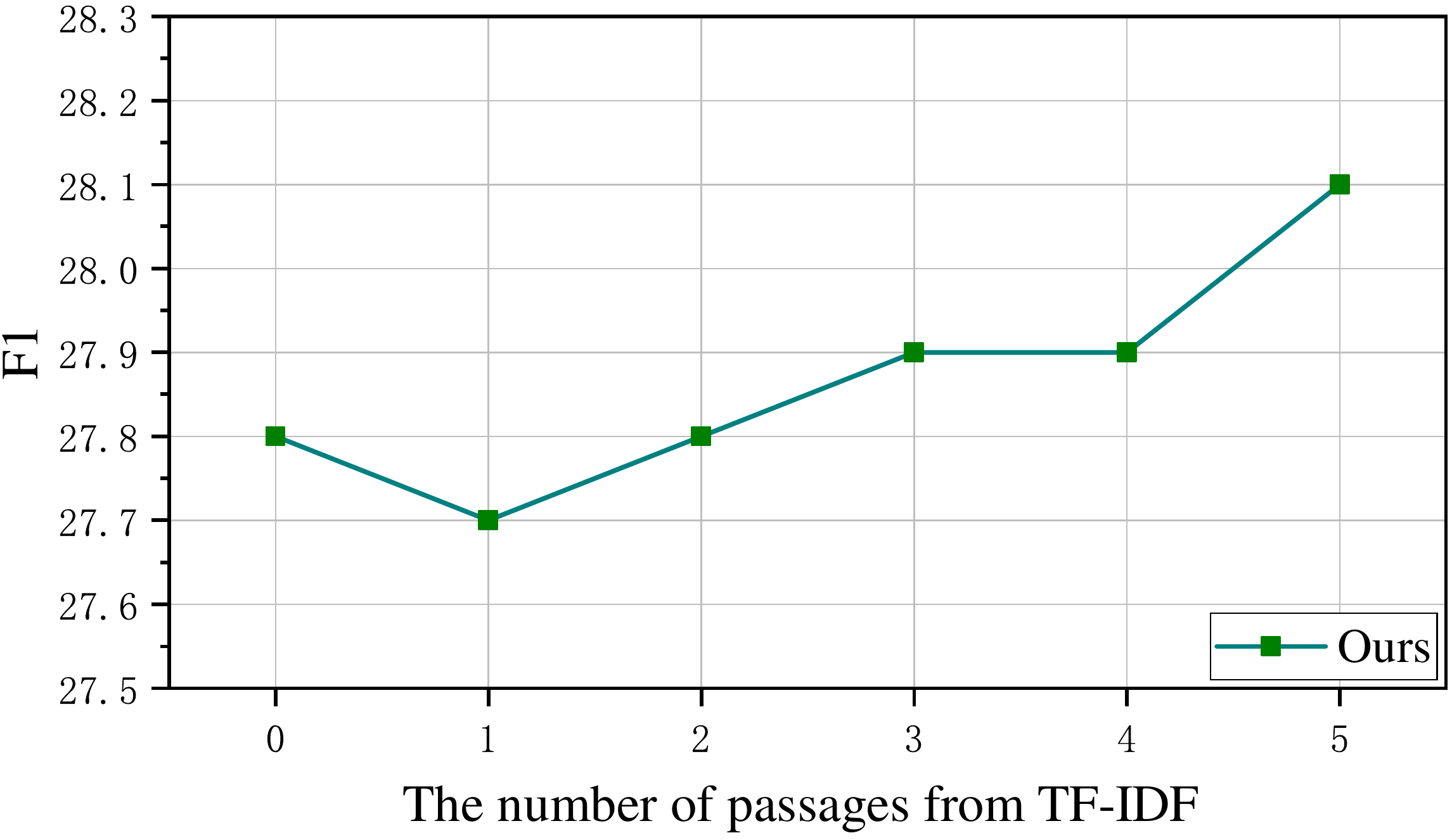}
   \label{fig6:subfig2}
   }
   \subfigure[]{

  \includegraphics[width=0.32\linewidth]{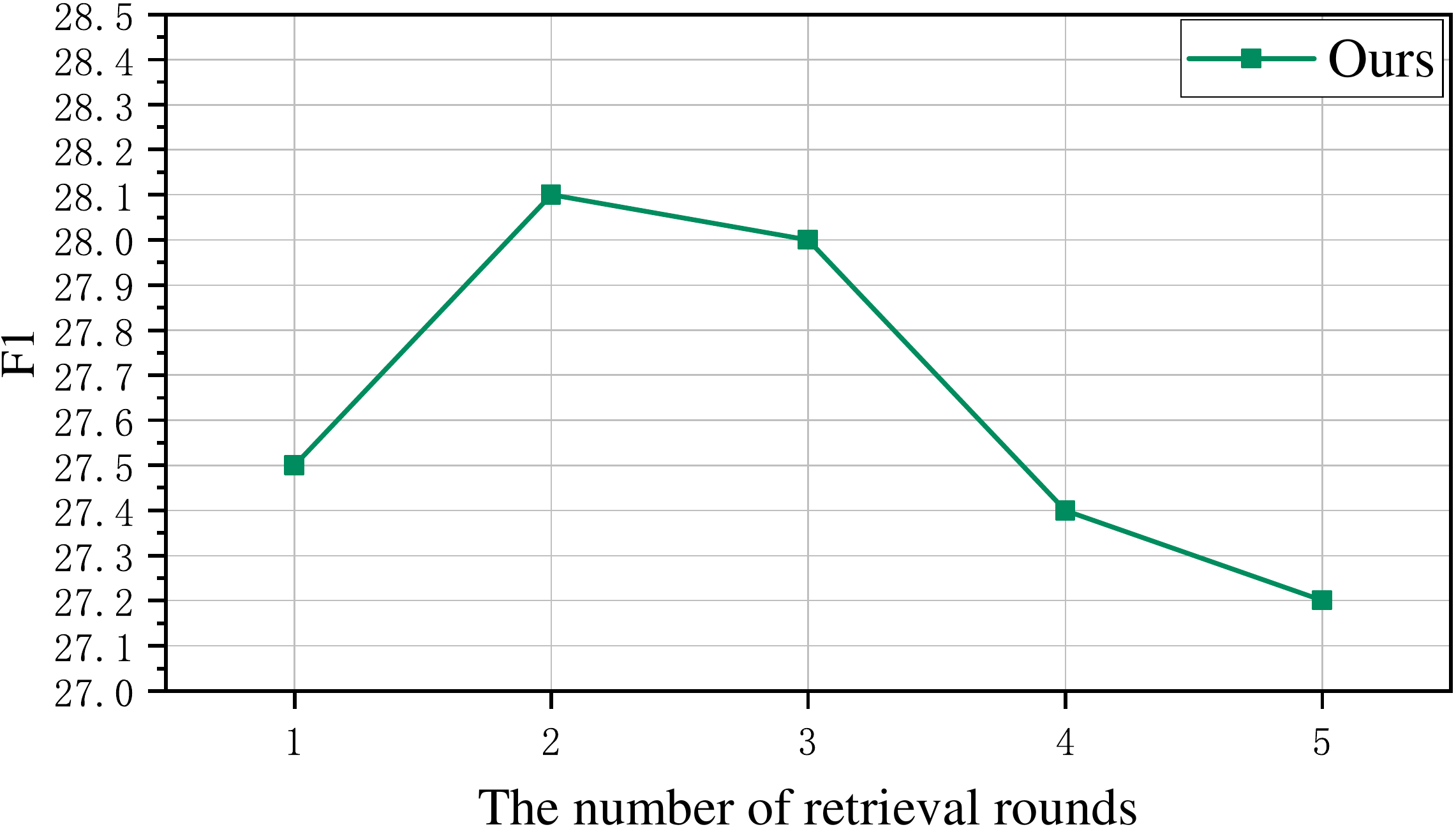}
   \label{fig6:subfig3}
   }
   \vspace{-1em}
  \caption{QA performance on the dev set versus the number of passages from the retriever, TF-IDF, and the number of the retrieve rounds, respectively. }\label{overall_fig}
   \vspace{-0.5em}

\end{figure*}

We compared these three variants, and summarized the results in Table~\ref{component_wise}.
By jointly analyzing Table~\ref{component_wise}, we gained the following insights.

(1) Removing the ranker component degrades the QA performance. To be more specific, ``Ours w/o Ranker'' drops by 0.3 in terms of F1. This statistic reveals the effectiveness of the ranker component. Although the ranker loss does not influence the retriever, the retriever performance also decreases. This is because that the joint training  mechanism can improve all different components. 

(2) ``Ours'' surpasses ``Ours w/o Explorer'', which indicates that incorporating the passage graph is indeed beneficialin boosting the performance. Moreover, compared with  ``Ours w/o Ranker'', the performance of ``Ours'' drops more, which again reflects the effectiveness of our graph-guided retrieval component. 

(3) ``Ours'' shows the consistent improvements over ``Ours w/o DHM''. The improvement in terms of F1 is 0.8. Such phenomenon clearly reflects the great advantage of our novel multi-round dynamic retrieval method.

(4) By comparing three variants, it is clear that when the Explorer is excluded, the system performance drops the most. We believed that the improvement of our method comes from the graph guided retrieval. Meanwhile, when the Ranker is excluded, it only causes less impact on the overall performance.

\begin{table*}[tbp]

\centering
\caption{Case study. The test samples of our proposed method. Attention weights refer to the attention score computed in Eqn.($\ref{eqn4}$). 1-round Retrieval, 2-round Retrieval, Explorer, and Ranker refer to the labels of the top-5 retrieved passages from the first round of Retrieval, the second round of Retrieval, the Explorer, and the Ranker, respectively.}\label{case study}
\vspace{-0.5em}
 \scalebox{0.95}{
\begin{tabular}{ccccccc}
\toprule
\multirow{2}{*}{Current question}&\multirow{2}{*}{History questions}&Attention &1-round&2-round&\multirow{2}{*}{Explorer}&\multirow{2}{*}{Ranker} \\
&&Weights&Retrieval& Retrieval&&\\
\midrule
\multirow{4}{*}{Did \textcolor{blue}{he} play any live shows?} 
&What led Hank Snow to Nashville?& 0.3781&\multirow{4}{*}{[1 0 0 0 0]} &\multirow{4}{*}{[1 0 0 0 0]} &\multirow{4}{*}{[0 0 1 0 0]}&\multirow{4}{*}{[1 0 0 0 0]} \\
&Where did he get his start in Nasville?& 0.2049&&&&\\
&What was his most popular song? &0.2313&&&&\\
&Did he win any awards &0.1858&&&&\\
\midrule
\multirow{2}{*}{Why did \textcolor{blue}{he} burn \textcolor{blue}{it}?} 
&Did Varg Vikernes commit any arson?&0.5521&\multirow{2}{*}{[0 0 0 0 0]} &\multirow{2}{*}{[0 0 0 0 0]} &\multirow{2}{*}{[0 0 0 0 1]}&\multirow{2}{*}{[1 0 0 0 0]} \\
&When was the first case?&0.4479&&&&\\

\midrule

Did \textcolor{blue}{he} accept and appear
&What was Sakis Rouvas' first&\multirow{2}{*}{1}&\multirow{2}{*}{[0 0 0 0 0]} &\multirow{2}{*}{[0 0 0 1 0]} &\multirow{2}{*}{[0 0 0 0 1]}&\multirow{2}{*}{[0 0 1 0 0]} \\
on \textcolor{blue}{the show}?&tv appearance?&&&&&\\

\bottomrule
\end{tabular}}

\end{table*}

\subsection{Additional Analyses}
\textbf{Impact of the number of passages in $\mathcal{P}_{retriever}$}. As mentioned in Section 3.2, the Retriever component retrieves $n_1$ passages and then Explorer utilizes these passages to activate the corresponding nodes in the passage graph. Therefore, $n_1$ is an import hyper-parameter that needs to be explored. To this end, we carried out experiments on the development set to verify the impact of the number of passages in $\mathcal{P}_{retriever}$. The choices of $n_1$ are from $0$ to $9$, and the comparison results  are illustrated in Figure~\ref{fig6:subfig1}. It demonstrates that if the passages from the Retriever are not included to activate the nodes, the performance drops a lot. This is because that the retrieval result help locate an appropriate area in the big passage graph, and thus it contributes to the retrieval of the gold passage.  Besides, as the number of passages from the Retriever increases, the F1 score rises and then gradually degrades. This experimental result tells us that more seed nodes do not represent better performance, because it also make it more difficult to retrieve the gold one when the number of nodes in the sub-graph gets larger.

\textbf{Impact of the number of passages from TF-IDF}. Similarly, the Explorer utilizes the passages from TF-IDF to activate the nodes in the passage graph. Therefore, we conducted experiments to explore  the number of passages from TF-IDF, and the number is set from $0$ to $5$. The results are summarized in Figure~\ref{fig6:subfig2}. We can observe that selecting a certain amount of passages from TF-IDF as seed nodes benefiting the QA system. However, the performance drops when the number of passages is increased from $0$ to $1$. This may be due to the fact that the quality of the passage list from TF-IDF is not as good as that from the dense retriever, thus the position of  the gold passage is not at the front. And as the number of passages increases, the F1 score gradually rises because more and more relevant passages are added. 

\textbf{Impact of the number of retrieve rounds}. As claimed before, our proposed DHM makes the multi-round retrieval possible, thus it is worth to exploit the number of retrieval rounds. However, due to the limitation of the GPU memory, we only tuned this value on the dev set when evaluating the system. The results are shown in Figure~\ref{fig6:subfig3}. We found that increasing the retrieval rounds does not improve the performance continuously. On the one hand, we did not tune the number of retrieval rounds in the training process. On the other hand, multi-round retrieval may bring more noise and probably suffer from the error propagation problem.

\subsection{Case study}
To gain deeper insights into the performance of our proposed method in conversational open-domain QA, we listed examples in Table~\ref{case study}. Each example includes the current question, history questions, and attention weights. To illustrate our proposed graph-guided and multi-round retrieval method, we also listed the labels of the retrieved passages from the 1-round retrieval, 2-round retrieval, the Explorer, and the Ranker. From examples in Table~\ref{case study}, we had the following observations.

\textbf{Current questions}. (1) Analyzing the current questions, we found that the coreference problem is common. For example, the pronoun ``he'' in question ``Did he play any live shows?'' refers to the name ``Hank Snow'' in the first history question. (2) It is also observed that there may be two coreferences in one question. For example, for the question ``Did he accept and appear on the show'', ``he'' and ``the show'' refer to ``Sakis Rouvas'' and ``the first tv appearance'' in the history question, respectively. (3) Moreover, the current question may depend on the history answers. For the second question ``Why did he burn it?'', we cannot clarify what the word ``it'' means only based on the history questions, and which turns out to be ``the Fantoft Stave Church'' in the previous answer ``On 6 June 1992, the Fantoft Stave Church, dating from the 12th century and considered architecturally significant, was burned.''. This observation is consistent with our experiment results that ``Ours-ta'' and ORConvQA-ta largely surpass the ones  than that without the previous true answers. 

\textbf{Attention weights}. (1) By jointly analyzing the history questions and attention weights, we found that the attention weights approximately reflect the importance of history questions. For example, in the first and second example, the first history question obtains the largest attention score. This adheres to our intuition because the first history question usually mentions the topic of the conversation. (2) It surprises us that the second history question of the second example also achieves a comparative attention score. As claimed before, in the second example, the gold passage and answer of the second history question provide important information to help understand the current question. Therefore, a comparative attention score becomes reasonable because it makes it easier to retrieve the corresponding evidence to clarify the current question. (3) The attention weights are not significantly discriminative. For example, in the first example, the attention weight of the fourth history question is 0.1858 although it may only provide little of useful information. This is partly due to the fact the questions in a conversation are always related more or less, since they usually have a same or similar topic.

\textbf{Retrieval result}. (1) From the result of the 1-round of retrieval, we found that it is easier for the dense retriever to handle the simple question, like the first example that contains one pronoun. However, for some more complex questions, it seems that it works not so well. (2) Analyzing the result of 2-round of retrieval, we found that it can make a supplement for the 1-round of retrieval in some cases. For example, in the third example, 1-round of retrieval does not retrieve the gold passage but 2-round of retrieval does. This may illustrate  our dynamically multi-round retrieval method works partly due to its ability of handling the complex questions. And this observation is consistent with some recent studies~\cite{xiong2020answering,das2018multi} that focus on applying the multi-round retrieval method  to answer the complex questions in open-domain QA. (3) It also can be observed that the Explorer obtains the best retrieval result, which is consistent with the experiment results in Section 4.4. For example, in the second example in Table~\ref{case study}, the Explorer retrieves the gold passage while both the 1-round and 2-round of  retrieval do not. This is because our proposed Explorer models the relation among answers in the conversation. However, the position of the gold passage in the result of the Explorer is not as front as in the result of the Retriever. This is may be because of  the graph neural network, but it is acceptable because the main target of retrieval is to improve the recall as much as possible. And fortunately, the following Ranker component refines the rank list.

\section{Conclusion and Future Work}
In this work, we present a novel conversational open-domain QA method, including DHM, Retriever, Explorer, Ranker, and Reader. Different from the previous work, our method focuses on exploiting the relations among answers across the conversation rather than only utilizing the historical context to enhance the current question. Specifically, we design the Explorer, which explores the neighbors of the passages from the Retriever, TF-IDF, and history answers via the passage graph. Moreover, we utilize the relevance feedback to dynamically attend the useful information of the historical context. Based on the dynamic question embeddings, we develop the multi-round retrieval method to retrieve more evidence for the subsequent answer extraction. To justify our method, we perform extensive experiments on the
public dataset, and the experimental results demonstrate the effectiveness of our model. 

In future, we plan to deepen and widen our work from the following aspects: (1) In this work, we utilized the hyperlinks in Wikipedia to construct the passage graph. Due to the practical concern that not all passages comes from Wikipedia, we will extend our method to the situation that the hyperlinks are hard to obtain by introducing the knowledge graphs.  (2) the OR-QuAC dataset is from the conversational MCR area. There are still some differences between conversational open-domain QA and conversational MCR.  Thus, it is necessary to conduct a dataset more suitable for the real conversational open-domain QA, where people ask questions for the purpose of their real information need. And (3), as shown in the Case Study, there are still complex questions in conversational QA. We will focus on these complex questions in our future work.

\bibliographystyle{ACM-Reference-Format}
\balance
\bibliography{acmart}

\appendix

\end{document}